# MediaGPT : A Large Language Model for Chinese Media

**Zhonghao Wang, Zijia Lu, Bo Jin, Haiying Deng**
State Key Laboratory of Media Convergence Production Technology and Systems, China
{wangzhonghao, luzijia, jinbo}@xinhua.org, denghaiying@xinhuaskl.com

## Abstract

Large language models (LLMs) have shown remarkable capabilities in generating high-quality text and making predictions based on large amounts of data, including the media domain. However, in practical applications, the differences between the media's use cases and the general-purpose applications of LLMs have become increasingly apparent, especially Chinese. This paper examines the unique characteristics of media-domain-specific LLMs compared to general LLMs, designed a diverse set of task instruction types to cater the specific requirements of the domain and constructed unique datasets that are tailored to the media domain. Based on these, we proposed MediaGPT, a domain-specific LLM for the Chinese media domain, training by domain-specific data and experts SFT data. By performing human experts evaluation and strong model evaluation on a validation set, this paper demonstrated that MediaGPT outperforms mainstream models on various Chinese media domain tasks and verifies the importance of domain data and domain-defined prompt types for building an effective domain-specific LLM.

***K*eywords**  Artificial Intelligence · Large Language Model · Media

## 1 Introduction

Large Language Models (LLMs) have gained significant attention in recent years for their ability to generate high-quality text and make predictions based on large amounts of data. OpenAI's InstructGPT research [1] laid the foundation for this technology, and subsequent products like chatGPT [2] have demonstrated its effectiveness in diverse fields such as natural language processing, machine translation, and content creation.

The potential of LLM technology has been increasingly realized by the public. However, in order for LLMs to be effectively applied in industries and assist in improving work efficiency, they need to be tailored to meet specific domain demands. There are two main approaches to enhancing LLMs for domain-specific applications. One approach is to use prompts to improve model accuracy [3], while another approach is to train the model using domain-specific data, such as BloombergGPT and FinGPT [4, 5].

In the media domain, for example, many professionals have found that generic LLMs often fail to meet their expectations due to the domain's particular characteristics, such as unique writing styles, narrative structures, and even differing political stances among media outlets [6, 7]. Consequently, there is a growing need for LLMs specifically designed for the media domain.

This paper takes the media domain as the entry point and highlights the uniqueness of domain-specific large models compared to general large models. These unique features are not only reflected in pre-training samples, but also in various aspects such as prompts, SFT data instructions, and verification methods. Therefore, MediaGPT, an domain-specific large model designed for the Chinese media domain is proposed in this paper. It is trained using domain-specific data and fine-tuned using experts SFT data that capture media business requirements. With the help of experts, we constructed a validation set for the Chinese media domain, and performed human experts evaluation and strong model evaluation on it. Ultimately, MediaGPT demonstrates superior performance in Chinese media domain tasks and verified the importance of domain data.

## 2 Related Work

Large language models (LLMs) are neural network models that are trained on massive amounts of text data, and can generate natural language for various tasks and domains. In this section, we review some of the related work on LLMs, focusing on two aspects: domain-specific large language models (DS-LLMs) and Chinese large language models (CLLMs). DS-LLMs are LLMs that are trained or adapted on data from a specific domain, such as biomedical, legal, or financial text. CLLMs are LLMs that are trained or adapted on data from the Chinese language, which is one of the most widely spoken and written languages in the world. We discuss the potential and challenges of developing and applying DS-LLMs and CLLMs for various domains and applications.

### 2.1 Domain-specific Large Language Models

Domain-specific large language models (DS-LLMs) are LLMs that are trained or adapted on data from a specific domain, such as biomedical, legal, or financial text. DS-LLMs aim to capture the domain knowledge, terminology, and style of the target domain, and improve the performance of various downstream tasks in that domain. DS-LLMs can be obtained by either training a new LLM from scratch on domain data, or fine-tuning a pre-trained LLM on domain data. The latter approach is a form of transfer learning, where the pre-trained LLM serves as a general-purpose model that provides a good initialization for the domain-specific model.

Several studies have shown the benefits of DS-LLMs for different domains and tasks. For example, BioMedLM is a 2.7B parameter GPT model trained on biomedical data from PubMed, which achieves state-of-the-art results on medical question answering [8]. [9] provides a systematic taxonomy and review of DS-LLM techniques based on the accessibility to LLMs and the availability of domain data. BloombergGPT is a large language model that is specifically designed for the financial domain, with 50 billion parameters and the ability to handle various financial data and tasks[4]. It is an improvement over Bloom model, using Bloomberg's rich data sources and general-purpose datasets for training. It performs well on standard language model benchmarks and open financial benchmarks, as well as some internal financial tasks. These works illustrate the potential and challenges of developing and applying DS-LLMs for various domains and applications.

### 2.2 Chinese Large Language Models

Chinese large language models (CLLMs) are LLMs that are trained or adapted on data from the Chinese language, which is one of the most widely spoken and written languages in the world. CLLMs aim to capture the linguistic and cultural diversity, complexity, and richness of the Chinese language, and improve the performance of various downstream tasks in Chinese natural language processing. CLLMs can be obtained by either training a new LLM from scratch on Chinese data, or fine-tuning a pre-trained LLM on Chinese data. The latter approach is a form of transfer learning, where the pre-trained LLM serves as a general-purpose model that provides a good initialization for the Chinese-specific model.

Several studies have shown the benefits of CLLMs for different domains and tasks. For example, Safety Assessment of Chinese Large Language Models[10] is a technical report that evaluates and analyzes 15 CLLMs including the OpenAI GPT series and other well-known Chinese LLMs, where they observe some interesting findings on their safety issues and challenges. [11] is a paper that presents a practice on training large-scale autoregressive language models named PanGu, with up to 200 billion parameters, which demonstrates superior capabilities of performing various tasks under few-shot or zero-shot settings. An open source CLLM named Wu Dao 2.0 was trained using FastMoEs by Beijing

Academy of Artificial Intelligence (BAAI), which claims to be 10 times larger than GPT-3 and can handle both natural language and images[12]. These works illustrate the potential and challenges of developing and applying CLLMs for various domains and applications.

## 3 Dataset

### 3.1 Unlabeled Pretrain Data

Unlabeled pretrain data is the data used for pre-training LLM before fine-tuning it on the target domain of Chinese media. We selected all the published data from influential Chinese media outlets since about 2000, as well as the published data from influential English media outlets with official backgrounds , totally about 250GB. These data sources are produced by highly professional media practitioners, and have high quality and credibility. The data provides a rich and diverse source of text data for LLM to learn from. It covers a wide range of topics and domains related to Chinese media, such as politics, economy, culture, society, sports, entertainment, science, technology, health, education, and international affairs, etc. It also includes some English texts from official media outlets that can help our LLM learn cross-lingual knowledge and skills. The unlabeled pretrain data can enhance the generality and robustness of the LLM's representations and knowledge for various tasks or domains in Chinese media.

### 3.2 Supervised Fine-tuning(SFT) Data

Supervised fine-tuning (SFT) is a technique that involves training or adapting a pre-trained large language model on a specific task or domain, using labeled data. SFT can be used to improve the performance of a language model on natural language generation, question answering, or text summarization tasks, by fine-tuning it on data that contains input-output pairs for the desired task or domain.

To enable MediaGPT to better learn the main tasks in the media domain, we surveyed the expert opinions of more than ten media companies or departments in China, and designed four major categories of SFT based on the main needs of Chinese media practitioners in the workflow of editing, writing and publishing, see Table 1. These categories contain over 80 specific types of SFT data:

- **Opinion creation**, which requires the MediaGPT to generate more content without deviating from the facts given by the prompt, such as adding opinions and presenting positions. Typical business scenarios include: outline generation based on topics, comment article generation, etc.

- **Article transcription**, which requires the MediaGPT to create content based on the information given by the prompt, without adding opinions or modifying facts, such as news script generation based on manuscripts, news style transcription, summary article writing, etc.

- **Media understanding**, which requires the MediaGPT to understand and extract the content or elements of the information given by the prompt, without adding opinions or modifying facts, such as headline generation, news element extraction, etc.

- **Other QA**, which are some additional general knowledge QA samples to improve the generalization and emergence abilities of the MediaGPT.

In the process of constructing these instruction samples, we aimed to make the MediaGPT's creation more professional. For the SFT samples that mainly focus on media creation, we used real-world excellent media articles as outputs, and let humans or code generate inputs. This way, the difficulty of constructing SFT samples was greatly reduced, and the professional outputs also improved the quality of the MediaGPT's generation.

Table 1: Rules and examples for constructing the Chinese media instruction dataset

| Major categories | Examples of specific instruction type | Example rules of specific samples {} *means referenced content* |
|---|---|---|
| Opinion creation | Theme-based review article writing | Q:请基于{ theme}主题生成一篇评论类文章<br>A:{published high-quality commentary news} |
| | Outline creation | Q:请基于以下内容生成一篇文章大纲:<br>  {theme, information or something}<br>A:{article outline} |
| Article transcription | Summary generation | Q:请你根据以下新闻生成一段100字左右综述,<br>省略与稿件无关的信息: {a published article}<br>A:{the abstract article that meets the requirements} |
| | New media style transcription | Q:请你将以下严肃文章转写为一篇公众号风格稿件:<br>  {a published article}<br>A:{corresponding article published on WeChat} |
| Media understanding | Headline generation | Q:请为下面文章生成一个权威媒体风格的标题:<br>  {a published article}<br>A:{headlines of the article in Q} |
| | News element extraction | Q:请提取下面文章的新闻要素:<br>  {a published article}<br>A:{5W1H of the article in Q} |
| Other QA | Self-awareness instruction | Q:你叫什么?<br>A:我叫MediaGPT |

## 4 Model

In recent times, it has come to our attention that LLMs have opened up new possibilities and sparked a revolution in various domains. The LLMs in media domain differ primarily in their application areas and the datasets they are trained on.This paper aims to improve the performance of LLMs in the media domain through the implementation of domain-specific pre-training and the precise definition of datasets for supervised fine-tuning. Additionally, we build upon the structure of popular generative open-source LLMs' structure to complete our model.

Initially, we employed two popular generative open-source LLMs: BLOOMZ-7B[13] and LLaMA-7B[14] respectively as our base models. However, as our experiments progressed, it became evident that LLaMA-7B consistently outperformed BLOOMZ-7B. Therefore, we propose MediaGPT, a novel approach that involves supervised fine-tuning of the pre-trained model: LLaMA-7B. This decision was driven by LLaMA's superior performance and suitability for media domain tasks.

### 4.1 Architecture

LLaMA is a collection of LLMs trained on publicly available datasets and achieving efficient performance on various benchmarks. A variety of model sizes were trained ranging from 7 billion to 65 billion parameters. The underlying code of LLaMa is made available to researchers, enabling them to adjust the model according to their needs for various research, without any requirement for commercial licenses. LLaMA is designed as a multi-functional model that is suitable for various use cases, rather than just fine-tuning models for specific tasks. Furthermore, its requirements for computational power are relatively low.

LLaMA is based on the transformer architecture with various improvements that were subsequently proposed. Compared to GPT-3[15], LLaMA incorporates RMSNorm as a normalizing function to enhance training stability by normalizing the input of each transformer sub-layer, replaces ReLU non-linearity with the SwiGLU activation function for improved performance, and replaces absolute positional embeddings with rotary positional embeddings (RoPE) added at each layer of the network.

### 4.2 Scale

MediaGPT is based on LLaMA-7B so the parameters of the model is about 7 billion. The number of hidden layers is 32, the number of attention heads is 32 and the hidden size is 4096. LLaMA possesses an impressive multilingual and cross-lingual understanding capability, especially within European languages. Its training dataset consists of 1.4 trillion tokens, primarily in English, along with some other European languages in Latin or Cyrillic scripts [14]. As a result, its ability to generate Chinese text is limited. To address this issue and improve encoding efficiency, we propose augmenting the LLaMA vocabulary with additional Chinese tokens and employing the model for the extended vocabulary[16]. By combining the Chinese tokenizer with the original LLaMA tokenizer, we have created the Chinese LLaMA tokenizer, which now encompasses about 50k tokenized words.

### 4.3 Training

In this paper, we present the MediaGPT model, which is built upon the open-source generative model LLaMA-7B. Initially, the model was pre-trained on unlabeled data from the Chinese media domain, as described in **Section 3.1**. Subsequently, we conducted fine-tuning using two different SFT datasets. The version fine-tuned on the open-source dataset is referred to MediaGPT-generalSFT, while the one fine-tuned on the Chinese media domain dataset described in **Section 3.2** is referred to MediaGPT-domainSFT.

## 5 Evaluation

Quantitative evaluation of generative large language models is challenging, as there is no single dimension that can directly measure their performance. Most objective methods use multiple-choice or judgment questions, but the main goal of the Chinese media large language model is to handle subjective and open-ended questions, such as writing and creativity. These kinds of questions are best evaluated by human judges, but this approach is time-consuming and costly. Some recent works have used strong models as judges[17], where the models select which model is better, which slightly reduces the evaluation cost, but also severely affects the credibility.

We collected opinions from dozens of media domain experts, and selected several typical cases for evaluation in the Chinese media scenario, including human evaluation and strong model evaluation. We developed domain-specific dataset for the evaluation task, which is about main categories mentioned in **Section 3.2**. Each main category contains about 100 questions. Answers were generated by four different LLMs: ChatGPT-3.5[2], ERNIE Bot[18], MediaGPT-generalSFT and MediaGPT-domainSFT. ChatGPT-3.5 and ERNIE Bot are mainstream English and Chinese 100B-level models respectively, and we use api for testing. It should be noted that about 2% of ernie's answers refused to answer because of some security issues. The architecture, scale and pre-training data of MediaGPT-generalSFT and MediaGPT-domainSFT are exactly the same. The only difference between them is the SFT data, as mentioned in **Section 4.3**.

We proposed three performance evaluation metrics to quantify the quality and relevance of the model's outputs for each case: Avg.rank, Rank-n rate, and Compared win rate. We calculated the Rank-n rate for each model, which is the number of times the model is ranked 1st, 2nd, 3rd, or 4th. Avg.rank is the average of Rank-n. Compared win rate is the winning probability under the situation of pairwise comparison.

### 5.1 Human experts evaluation

In human evaluation, we enlisted the participation of a dozen journalists and editors as media domain experts to assess the results of the random half evaluation set, which answers generated by four different LLMs.

Before the evaluation process, we firstly asked some of the most senior of the dozen experts to devise evaluation criteria based on the news values and editorial standards. For example, for Opinion Creation tasks, the criteria encompassed five general evaluation dimensions and specific dimensions for each subtask. As for Article Transcription tasks, the criteria primarily comprised four general evaluation dimensions and specific dimensions for respective subtask.

In the formal evaluation process, we formulated a series of questions for all types of tasks and had the ChatGPT-3.5, ERNIE Bot, MediaGPT-generalSFT and MediaGPT-domainSFT generate answers to these questions. The media domain experts conducted a double-blind test of the answers provided by the four LLMs, and throughout this process, the experts were required to assign scores and rank the answers strictly following predetermined criteria.

The human evaluation results are as follows ,see Table 2. It can be observed that under the criteria of news expertise, MediaGPT-domainSFT achieves the highest average rate and the Rank-1 rate, and the win-rate. MediaGPT-domainSFT performs well in the following two aspects: 1) Understanding the question within the realm of news and consistently crafting the answers in the role of mainstream media. 2) Producing outputs that align better with the format of news, avoiding excessive stiffness. However, MediaGPT-domainSFT still has some shortcomings, such as sometimes missing the details of the question and writing in an overconceptualization way.

Known as main stream models, ChatGPT-3.5 and ERNIE Bot's most notable features lie in their meticulous comprehension of the nuances of the questions and clear presentation logic. But it's like a double-edged sword, ChatGPT-3.5 and ERNIE Bot tend to be overly formulaic in their logic and expressions, and they sometimes fail to contextualize certain abstract topics within the domain of news and current affairs. Besides, ERNIE Bot falls short compared to ChatGPT-3.5 in terms of shorter and less logical answer, and it sometimes tends to avoid addressing sensitive political topics, which is critical for news writing and may not meet the necessary usage requirements.

Regarding MediaGPT-general SFT, its biggest problem is often generating results in the wrong format. This indicates the LLM couldn't tell the difference of each news genre and emphasizes the significance of domain SFT.

Table 2: Results of human experts evaluation

| Model name | Avg. rank | Rank-n rate | | | | Compared win rate | | | |
|---|---|---|---|---|---|---|---|---|---|
| | | Rank-1 | Rank-2 | Rank-3 | Rank-4 | ChatGPT-3.5 | ERNIE-Bot | MediaGPT-generalSFT | MediaGPT-domainSFT |
| ChatGPT-3.5[2] | 2.19 | 29.9% | 35.8% | 19.4% | 14.9% | - | **65.7%** | 70.1% | **44.8%** |
| ERNIE-Bot[18] | 2.60 | 14.9% | 32.8% | 29.9% | 22.4% | 34.3% | - | 68.7% | 37.3% |
| MediaGPT-generalSFT | 3.18 | 10.5% | 13.4% | 23.9% | 52.2% | 29.9% | 31.3% | - | 20.9% |
| MediaGPT-domainSFT | **2.03** | **44.8%** | 17.9% | 26.9% | 10.5% | **55.2%** | 62.7% | **79.1%** | - |

## 5.2 Strong model evaluation

Conventional reference-based metrics, such as BLEU and ROUGE, have demonstrated relatively low correlation with human judgments. [19] discussed the potential of employing a more strong model to evaluate the performance of less strong models. Additionally, [20] delved into methods for reducing bias and increasing the alignment with human judgment in the strong models evaluation.

We use GPT-4 as the evaluation model and have proposed an evaluator prompt set for media-domain LLMs. Based on [21], strong model evaluation has the problem of validity and reliability. We present our Finetuned Evaluator Prompts Set, using four techniques to improve the stability and precision of evaluation. 1) Insert empty answer text in the first position of the answer list generated by LLMs given to the evaluation model to prevent evaluation biases caused by the order of the answers. 2) Randomly mix the order of the other answers for the same reason mentioned above. 3)

Reference each answer with normal names to avoid chaos caused by the additional meaning of the reference name. 4) Rank the answers instead of scoring them to minimize the instability caused by the evaluation model. Through our experiments, we found that by using finely-tuned prompts, the accuracy and stability of the evaluation were significantly improved.

The evaluator prompt set is the same as human experts evaluation. We monitored the performance of our models MediaGPT-generalSFT and MediaGPT-domainSFT with two other mainstream models ChatGPT-3.5 and ERNIE Bot. The results demonstrate the MediaGPT's superior performance in Chinese media domain tasks. The comparison of MediaGPT-generalSFT and MediaGPT-domainSFT proves that the performance of DS-LLMs improves steadily, and correlates with the using of SFT data, see Table 3.

In our experiments, we saw that finetuned evaluator prompts can enable strong model evaluation to achieve similar results to human experts judgement. The final judgement results 3 also shows a great similarity with those of humans 2, which further demonstrated the value of prompt tuning. However, it cannot be denied that the unstable performance of Strong Model itself [22] still poses a certain risk to the use of Strong model evaluation, even though it has the advantage of being fast and scalable.

Table 3: Results of strong model (GPT-4) evaluation

| Model name | Avg. rank | Rank-n rate | | | | Compared win rate | | | |
|---|---|---|---|---|---|---|---|---|---|
| | | Rank-1 | Rank-2 | Rank-3 | Rank-4 | ChatGPT-3.5 | ERNIE-Bot | MediaGPT-generalSFT | MediaGPT-domainSFT |
| ChatGPT-3.5[2] | 2.62 | 6.7% | 46.2% | 25.2% | 21.9% | - | 38.7% | 60.5% | **38.7%** |
| ERNIE-Bot[18] | 2.37 | 24.4% | 24.4% | 41.2% | 10.1% | 61.3% | - | 63.0% | 38.7% |
| MediaGPT-generalSFT | 2.87 | 14.3% | 24.4% | 21.9% | 39.5% | 39.5% | 37.0% | - | 37.0% |
| MediaGPT-domainSFT | **2.14** | **54.6%** | 5.0% | 11.8% | 28.6% | **61.3%** | **61.3%** | **63.0%** | - |

### 5.3 Some results

In the appendix 4, we show some typical questions in the working scenarios of the Chinese media domain, as well as the answers given by MediaGPT.

## 6 Conclusion

In this paper, we presented MediaGPT, a large language model for the Chinese media domain, which can generate high-quality and relevant outputs for various tasks in the Chinese media domain. We propose and construct a set of media instructions to cover the main demands of the Chinese media workflow, and built a system of SFT data based on these. Due to the difficulty of evaluation in the media domain, we proposed a adversarial evaluation method that combines human evaluation and strong model evaluation to measure the quality of the generative tasks. Under this evaluation framework, we demonstrated the superior performance of MediaGPT over existing models on main media domain cases, and verified the importance of domain data and domain-defined prompt types for building an effective domain-specific large language models. We hope that MediaGPT can be a valuable resource and tool for Chinese media practitioners and researchers, and inspire more innovations and applications in this domain.

# A  Some results

Table 4: Some opinion generation results

| Questions by human | Answers by MediaGPT |
| --- | --- |
| 请你以"美食带动旅游"为主题写一个中短篇幅评论文章。 | 美食是旅游业的重要组成部分，它不仅能够吸引游客的眼球，还能够提升当地形象和知名度。随着国内经济的发展以及居民生活水平的提高，人们对于文化、艺术等精神层面的需求也日益增长。在推动经济发展的同时，如何促进文化传承与发展成为当下亟待解决的问题之一。而通过发挥本土特色美食的优势来带动旅游产业，无疑是一种可行的方式。　　首先，美食可以为旅游产业带来流量。众所周知，餐饮业是最基础也是最有活力的经济行业之一，其对地方经济发展和社会就业的贡献不可小觑。因此，许多地方政府都把招商引资作为发展的重要手段，尤其是针对餐饮企业进行招商引资。这也就意味着，只要有好的餐饮项目落地，那么必然会带动大量的人流、车流及物流，进而带动相关产业链的发展。从这个角度来看，美食对于旅游产业发展的作用是不言而喻的。　　其次，美食也能够提升城市形象。每个地区都有自己独特的饮食文化，这些食物不仅是人们日常的生活必需品，更是体现地域文化底蕴的一种方式。例如，四川火锅、广东早茶、北京烤鸭等等，它们不仅仅是一道道美味佳肴，更代表着不同地区的历史积淀和文化传统。因此，将本地特色美食融入到城市形象建设之中去，既能展示出当地独有的风土人情，也可以让更多的游客了解并爱上当地的文化风情。然而，要想真正实现美食与旅游融合，还需要进一步完善配套服务。比如，打造具有特色的美食街区或美食市场；提供更加便捷高效的餐饮配送服务；开展相关的美食体验活动等等。只有这样才能让人们在品尝美食的同时，更好地感受当地的文化氛围。　　综上所述，美食与旅游有着紧密的联系，二者相辅相成的关系也在不断加深。未来，我们需要继续探索更多适合各地特色的美食元素，并将其充分植入到旅游活动中，从而形成更为丰富多彩的旅游产品，助力旅游业的发展。 |
| 你需要写一篇800-1000字的时事评论，行文要具备新华社风格，以【拒绝过度包装，让粽子回归"原味"】为标题，并依据这个新闻背景事件【层层拆开偌大一个礼盒，仅8个60克重的"迷你粽"，售价368元……端午临近，类似包装豪华、噱头十足、价格动辄数百上千元的粽子礼盒，陆续现身一些电商平台和大型商超、星级酒店等场所。】展开。 | 层层拆开偌大一个礼盒，仅8个60克重的"迷你粽"，售价368元……端午节临近，类似包装豪华、噱头十足、价格动辄数百上千元的粽子礼盒，陆续现身一些电商平台和大型商超、星级酒店等场所。　　一段时间以来，粽子成了商家眼中的"香饽饽"。从月饼到茶叶，再到如今的粽子礼盒，不断攀升的价格背后是愈演愈烈的铺张浪费风气。一些人为了追求高档次而斥巨资打造粽子品牌；一些人热衷于购买天价月饼送礼，转手倒卖赚取差价；还有人专门雇人在网上发布虚假信息，谎称降价出售高端粽子……凡此种种，不仅严重败坏了社会风气，更助长了"奢靡之始，危亡之渐"的不良倾向。　　"粽子经济"是市场经济的产物之一，原本用来纪念爱国诗人屈原。然而近年来，在商业利益驱动下，粽子逐渐变了味道：有人吃粽子却不买肉馅儿；有的人把粽子当零食，不加节制地食用甚至赌博式消费；更有甚者，将粽子作为营销手段，借机炒作概念进行"饥饿营销""拜金主义"……种种乱象导致粽子偏离了本义，背离了初心。　　从古至今，无论是清明上河图中包着粽子游街示众，还是端午节挂艾叶菖蒲祛病防疫，抑或是百姓自家包粽子祭奠先祖，都蕴含着中华民族对美好生活的向往与期盼。然而，随着物质生活日益丰富，人们越来越注重吃穿住行的品质提升，粽子这种"小家碧玉"渐渐被冷落。这其中固然有传统文化日渐式微的因素，但也不乏资本逐利和"唯利是图"的功劳。商家看准了人们对美好生活的向往和对传统文化的怀念，就千方百计迎合消费者心理，推出包装精美、价格高昂的粽子礼盒，借此满足部分人的虚荣心和猎奇心理，从而形成一种畸形的"粽子经济"。　　要刹住"粽子经济"歪风邪气，就要坚决摒弃畸形审美观，坚持一切从实际出发，量力而行、量入为出，避免铺张浪费。同时，有关部门应加强监管力度，严厉打击投机钻营行为，对"粽子经济"中的违规违法行为严惩不贷；同时还应倡导健康文明的饮食习惯，引导广大群众理性消费，自觉抵制低俗庸俗媚俗现象，共同营造崇俭抑奢的社会氛围。　　只有拒绝"粽子经济"过度包装，才能真正让粽子回归"原味"，实现真正的绿色消费、文明消费。 |

Table 5: Some article transcription results

| Questions by human | Answers by MediaGPT |
| --- | --- |
| 请对以下段落进行润色和改进，使语句通顺，用词精确：这几年，民政部一直在加强乡村地名建设的信息服务，关注到老百姓出行或者送快递找不到地方的现实问题，民政部让互联网地图平台按规范标注乡村地名670多万条，新增加的地名覆盖了全国200多个比较多人住的农村地方，这些新增的地名累计点击量有几十亿人次，这很好地促进了农村人民出行导航、快递送到村里和农产品送到城里。 | 近年来，民政部致力于改善农村地区居民的生活，在加强乡村地名信息服务和解决群众行路难、寄递不方便等问题方面进行了积极探索。为满足广大农民群众日益增长的出行需求和信息服务需要，民政部推动互联网地图平合按照国家标准规范标注乡村地名，共计标注地名670余万个，涵盖了全国200多个地区较为集中的人口区域。此举不仅使人们能够便捷地在手机等移动终端上获取相关信息，还有效解决了偏远地区的交通指引难题，受到了广大农民朋友的欢迎和支持。 |
| 请将严肃新闻报道改写为新媒体体裁：2023年中国无人机竞速公开赛（海南自贸港站）于7月13日-15日在海南省海口市举行。据组委会介绍，65名参赛选手中超半数都是青少年飞手，最小的参赛选手年仅6岁。作为一项新兴的智能科技运动，无人机竞速越来越受到青少年和家长的欢迎。现场不少选手家长表示，练习无人机竞速不仅丰富了孩子的课余生活，更提升了孩子的动手能力、抗压能力和独立解决问题的能力。除了专业赛事以外，本次比赛还设置了赛道体验、飞行游园会、迷你无人机竞速挑战赛等一系列丰富的周边活动。赛会主办方表示，选择在暑期举办这样的国家级无人机比赛，是希望在丰富孩子们的暑期生活的同时，鼓励更多青少年积极地参与到航空航天科技这样的未来科技的学习中来。 | 2023全国无人机竞速公开赛（海南自贸港站）<br>7月13日至15日在海口市举行<br>来自全国各地65位参赛选手<br>其中年龄最大的18岁<br>最小的只有6岁<br>经过两天的激烈比拼<br>"少年强则国强"<br>少年们展现出非凡的操控技术<br>现场不少选手家长说<br>通过练习无人机竞速<br>孩子不仅变得勇敢果断<br>还能培养他们的耐心和毅力<br>学好无人机这项技能<br>对孩子们来说很有意义<br>近年来<br>随着科技的发展与普及<br>越来越多的人接触到了无人机<br>并喜欢上了它<br>很多人认为<br>这项既刺激又酷炫的运动<br>将成为他们未来的日常<br>热爱科技运动的少年们<br>快来一起飞上蓝天吧！ |